\documentclass[journal]{IEEEtran}
\ifCLASSINFOpdf
\else
\fi
\usepackage{cite}
\usepackage[numbers]{natbib}
\usepackage{graphicx}  
\usepackage{subfigure}
\usepackage{amsmath}
\usepackage{booktabs}  
\usepackage{multirow}
\usepackage{CJK}
\usepackage{amsfonts}
\usepackage{url}
\usepackage{color}
\usepackage{caption}

\begin{document}
	%
	\title{Effective Subword Segmentation for \\ Text Comprehension}
	%
	%
	%
	
	\author{Zhuosheng Zhang, Hai Zhao, Kangwei Ling, Jiangtong Li, Zuchao Li, Shexia He, Guohong Fu
		
		\thanks{Manuscript received November 10, 2018; revised March 24, 2019; accepted June 04, 2019. This paper was partially supported by National Key Research and Development Program of China (No. 2017YFB0304100) and Key Projects of National Natural Science Foundation of China (U1836222 and 61733011). The associate editor coordinating the review of this manuscript and approving it for publication was Dr. Min Zhang (Corresponding author: Hai Zhao.).}
		\thanks{Zhuosheng Zhang, Hai Zhao, Kangwei Ling, Jiangtong Li, Zuchao Li, Shexia He are with the Department
			of Computer Science and Engineering, Shanghai Jiao Tong University, 800 Dongchuan Road, Minhang District, Shanghai, China. (e-mail: zhangzs@sjtu.edu.cn; zhaohai@cs.sjtu.edu.cn; kevinling@sjtu.edu.cn; keep\_moving-lee@sjtu.edu.cn; charlee@sjtu.edu.cn; heshexia@sjtu.edu.cn).}
	   	\thanks{Guohong Fu is with the Institute of Artificial Intelligence, Soochow University, China (email: ghfu@hotmail.com).}
		\thanks{Part of this study has been published as \emph{Subword-augmented Embedding for Cloze Reading Comprehension} \cite{Zhang2018Subword} in COLING-2018. This paper extends the previous byte pair encoding (BPE) subword method to introduce a unified segmentation framework, and conducts comprehensive experiments on both reading comprehension and textual entailment tasks, considering multilingual effectiveness, generalization ability on different benchmarks and thorough case studies. The codes have been released at https://github.com/cooelf/subword\_seg.}
	}
	
	%
	%

	\markboth{IEEE/ACM TRANSACTIONS ON AUDIO, SPEECH, AND LANGUAGE PROCESSIN, 2019}%
	{}
	
	%


	
    \IEEEoverridecommandlockouts
    \IEEEpubid{\begin{minipage}[t]{\textwidth}\ \\[10pt]
            \centering\normalsize{\copyright 2019 IEEE. Personal use is permitted, but republication/redistribution requires IEEE permission. \\ See http://www.ieee.org/publicationsstandards/publications/rights/index.html for more information.}
    \end{minipage}}

	\maketitle

	\begin{abstract}
		Representation learning is the foundation of machine reading comprehension and inference. In state-of-the-art models, character-level representations have been broadly adopted to alleviate the problem of effectively representing rare or complex words. However, character itself is not a natural minimal linguistic unit for representation or word embedding composing due to ignoring the linguistic coherence of consecutive characters inside word. This paper presents a general subword-augmented embedding framework for learning and composing computationally-derived subword-level representations. We survey a series of unsupervised segmentation methods for subword acquisition and different subword-augmented strategies for text understanding, showing that subword-augmented embedding significantly improves our baselines in various types of text understanding tasks on both English and Chinese benchmarks.
	\end{abstract}
	

	%
	\IEEEpeerreviewmaketitle

	\section{Introduction}
	
	The fundamental part of deep learning methods applied to natural language processing (NLP), distributed word representation, namely, \emph{word embedding}, provides a basic solution to text representation for NLP tasks and has proven useful in various applications, including textual entailment \cite{Wang2017Bilateral,zhang2018know} and machine reading comprehension (MRC) \cite{Seo2016Bidirectional,zhang2018OneShot,clark2018simple,zhang2019dual}. However, deep learning based NLP models usually suffer from rare and out-of-vocabulary (OOV) word representation \cite{Sennrich2015Neural,luong2016achieving}, especially for low-resource languages. 
	Besides, most word embedding approaches treat word forms as atomic units, which is spoiled by many words that actually have a  complex internal structure. Especially, rare words like morphologically complex words and named entities, are often expressed poorly due to data sparsity. Actually, plenty of words share some conjunct written units, such as morphemes, stems and affixes. The models would benefit a lot from distilling these salient units automatically.
	
	Character-level embedding has been broadly used to refine the word representation  \cite{Yang2016Words,Kim2015Character,luong2016achieving,Li2018Neural}, showing beneficially complementary to word representations. Concretely, each word is split into a sequence of characters. Character representations are obtained by applying neural networks on the character sequence of the word, and their hidden states form the representation. 
	
	However, character is not the natural minimum linguistic unit, which makes it quite valuable to explore the potential unit (subword) between character and word to model sub-word morphologies or lexical semantics. For English, there are only 26 letters. Using such a small character vocabulary to form the word representations could be too insufficient and coarse. Even for a language like Chinese with a large set of characters (typically, thousands of), lots of which are semantically ambiguous, using character embedding below the word-level to build the word representations would not be accurate enough, either. For example, for an internet neologism \begin{CJK*}{UTF8}{gkai}{老司机}\end{CJK*} (\emph{experienced driver}), the characters \begin{CJK*}{UTF8}{gkai}{$<$老 (\emph{experienced, old}) , 司 (\emph{manage}), 机 (\emph{machine})$>$}\end{CJK*} would be somewhat from the meaning of the word while the subwords \begin{CJK*}{UTF8}{gkai}{$<$老 (\emph{experienced, old}), 司机  (\emph{driver})$>$}\end{CJK*} with proper syntactic and semantic decomposition give exactly the minimal meaningful units below the word-level which surely improve the later word representation. Thus, in either type of languages, effective representation cannot be done accurately only via the character based process.
	
	In fact, morphological compounding (e.g. \emph{sunshine} or \emph{playground}) is one of the most common and productive methods of word formation across human languages, and most of rare or OOV words can be segmented into meaningful fine-grained subword units for accurate learning and representation, which inspires us to represent word by meaningful sub-word units. Recently, researchers have started to work on morphologically informed word representations \cite{Hammarstr2011Unsupervised,Botha2014Compositional,Cao2016A,Bergmanis2017From}, aiming at better capturing syntactic, lexical and morphological information. With flexible subwords from either source, we do not necessarily need to work with characters, and segmentation could be stopped at the subword-level. With related characters grouping into subword, we hopefully reach a meaningful minimal representation unit. 
	
	Splitting a word into sub-word level subwords and using these subwords to augment the word representation may recover the lost syntactic or semantic information that is supposed to be delivered by subwords. For example, \emph{understanding} could be split into the following subwords: $<$under, stand, ing$>$. Previous work usually considered prior linguistic knowledge based methods to tokenize each word into subwords (namely, \emph{morphological based subword}). However, such treatment may encounter two main inconveniences. First, the linguistic knowledge resulting subwords, typically, morphological suffix, prefix or stem, may not be suitable for the targeted NLP tasks. Second, linguistic knowledge or related annotated lexicons or corpora even may not be available for a specific language or task. Thus in this work we consider computationally motivated subword tokenization approaches instead. 
	
	We present a unified representation learning framework to sub-word level information enhanced text understanding and survey various computationally motivated segmentation methods. Specifically, we consider the subword as the basic unit in our models and manipulate the neural architecture accordingly. The proposed method takes variable-length subwords segmented by unsupervised segmentation measures, without relying on any predefined linguistic resource. First, a goodness score is computed for each $n$-gram using the selected goodness measure to form a dictionary. Then segmentation or decoding method is applied to tokenize words into subwords based on the dictionary. The proposed subword-augmented embedding will be evaluated on text understanding tasks, including textual entailment and machine reading comprehension, both of which are quite challenging due to the need of accurate lexical-level representation. Furthermore, we empirically survey various subword segmentation methods from a computational perspective and investigate the better way to enhance the tasks with thoughtful analysis and case studies.
	
	The rest of this paper is organized as follows. The next section reviews the related work. Section 3 will demonstrate our subword augmented learning framework and implementation. Task details and experimental results are reported in Section 4, followed by case studies and analysis in Section 5 and conclusion in Section 6.

	\begin{table}
		\centering
		\caption{\label{tab:mrc_example} A machine reading comprehension example.}
		{
			\begin{tabular}{l|p{6.5cm}}
				\hline
				\hline
				{\bf Passage }
				& 
				Robotics is an interdisciplinary branch of engineering and science that includes mechanical engineering, electrical engineering, computer science, and others. Robotics deals with the design, construction, operation, and use of robots, as well as computer systems for their control, sensory feedback, and information processing. These technologies are used to develop machines that can substitute for humans. Robots can be used in any situation and for any purpose, but today many are used in dangerous environments (including bomb detection and de-activation), manufacturing processes, or where humans cannot survive. Robots can take on any form but some are made to resemble humans in appearance. This is said to help in the acceptance of a robot in certain replicative behaviors usually performed by people. Such robots attempt to  replicate walking, lifting, speech, cognition, and basically anything a human can do.\\ 
				\hline
				{\bf Question} & What do robots that resemble humans attempt to do? \\
				\hline
				{\bf Answer} & replicate walking, lifting, speech, cognition \\
				\hline
				\hline
			\end{tabular}
		}
		
	\end{table}

	\begin{table}
		\centering
		\caption{\label{tab:rte_example} A textual entailment example.}
		{
			\begin{tabular}{l|l|l}
				\hline
				\hline
				\bf Premise & Man grilling fish on barbecue & Label\\
				\hline
				\multirow{3}{*}{\bf Hypothesis} & The man is cooking fish. & Entailment \\
				& The man is sailing a boat. &  Contradiction \\
				&  The man likes to eat fish. & Neutral \\
				\hline
				\hline
			\end{tabular}
		}
		
	\end{table}
	
	\section{Related Work}
	\subsection{Augmented Embedding}
	To model texts into vector space, the input tokens are represented as embeddings in deep learning models \cite{wang-etal-2017-sentence,8360031,zhang2019acl,zhang2018DUA,li2019dependency,li-etal-2018-seq2seq,li2018unified}. Previous work has shown that word representations in NLP tasks can benefit from character-level models, which aim at learning language representations directly from characters. Character-level features have been widely used in language modeling \cite{Miyamoto2016Gated,Peters2018ELMO}, machine translation \cite{luong2016achieving,Sennrich2015Neural} and reading comprehension \cite{Yang2016Words,Seo2016Bidirectional}.
	\citet{Seo2016Bidirectional} concatenated the character and word embedding to feed a two-layer Highway Network. \citet{Cai2017Fast} presented a greedy neural word segmenter to balance word and character embeddings. High-frequency word embeddings are attached to character embedding via average pooling while low-frequency words are represented as character embedding. \citet{Miyamoto2016Gated} introduced a recurrent neural network language model with LSTM units and a word-character gate to adaptively find the optimal mixture of the character-level and word-level inputs. \citet{Yang2016Words} explored a fine-grained gating mechanism to dynamically combine word-level and character-level representations based on properties of the words (e.g. named entity and part-of-speech tags). 
	
	However, character embeddings only show marginal improvement due to a lack of internal semantics. Recently, many techniques were proposed to enrich word representations with sub-word information. \citet{Bojanowski2016Enriching} proposed to learn representations for character $n$-gram vectors and represent words as the sum of the $n$-gram vectors. \citet{Avraham2017The} built a model inspired by \citet{Joulin2016Bag}, who used morphological tags instead of $n$-grams. They jointly trained
	their morphological and semantic embeddings, implicitly assuming that morphological and semantic information should live in the same space. Our work departs from previous ones on morphologically-driven embeddings by focusing on embedding data-driven subwords. To handle rare words, \citet{Sennrich2015Neural} introduced the byte pair encoding (BPE) compression algorithm for open-vocabulary neural machine translation by encoding rare and unknown words as subword units. \citet{Zhang2018Subword} applied BPE for cloze-style reading comprehension to handle OOV issues. Different from the motivation of sub-word segmentation for rare words modeling, our proposed unified subword-augmented embedding framework serves for a general purpose without relying on any predefined linguistic resources with thorough analysis, which can be adopted to the enhance the representation for each word by adaptively altering the segmentation granularity in multiple NLP tasks.

	\subsection{Text Comprehension}

	As a challenging task in NLP, text comprehension aims to read and comprehend a given text, and then answer questions or make inference based on it. These tasks require a comprehensive understanding of natural languages and the ability to do further inference and reasoning. In this paper, we focus on two types of text comprehension, document-based question-answering (Table \ref{tab:mrc_example}) and textual entailment (Table \ref{tab:rte_example}), which share the similar genre of machine reading comprehension, though the task formations are slightly different. 
	
	In the last decade, the MRC tasks have evolved from the early cloze-style test \cite{hill2015goldilocks,hermann2015teaching,Zhang2018Subword} to span-based answer extraction from passage \cite{Rajpurkar2016SQuAD,Rajpurkar2018Know,Nishida:2018:RML:3269206.3271702}. The former has restrictions that each answer should be a single word  in the document and the original sentence without the answer part is taken as the query. For the span-based one, the query is formed as questions in natural language whose answers are spans of texts. Notably, \citet{Chen2016A} conducted an in-depth and thoughtful examination on the comprehension task based on an attentive neural network and an entity-centric classifier with a careful analysis based on handful features. Then, various attentive models have been employed for text representation and relation discovery, including Attention Sum Reader \cite{kadlec2016text}, Gated attention Reader \cite{Dhingra2017Gated}, Self-matching Network \cite{Wang2017Gated} and Attended over Attention Reader \cite{Cui2016Attention}.
	
	With the release of the large-scale span-based datasets \cite{Rajpurkar2016SQuAD,Nguyen2016MS,Joshi2017TriviaQA,Yizhong2018Multi,Rajpurkar2018Know}, which constrain answers to all possible text spans within the reference document, researchers are investigating the models with more logical reasoning and content understanding \cite{Yizhong2018Multi,Wang2017Gated}.
	
	For the other type of text comprehension, natural language inference (NLI) is proposed to serve as a benchmark for natural language understanding and inference, which is also known as recognizing textual entailment (RTE). In this task, a model is presented with a pair of sentences and asked to judge the relationship between their meanings, including entailment, neutral and contradiction. \citet{Bowman2015A} released Stanford Natural language Inference (SNLI) dataset, which is a high-quality and large-scale benchmark, thus inspiring various significant work. 
	
	Most of existing NLI models apply attention  mechanism  to  jointly  interpret  and  align the premise and hypothesis,  while  transfer  learning  from
	external knowledge is popular recently. Notably, \citet{Chen2017Enhanced} proposed  an  enhanced  sequential  inference model (ESIM), which employed recursive architectures in both local inference modeling and
	inference composition, as well as syntactic
	parsing information, for a sequential inference model. ESIM is simple with satisfactory performance, and is thus widely chosen as the baseline model. \citet{Mccann2017Learned} proposed  to
	transfer  the  LSTM  encoder  from  the  neural
	machine translation (NMT) to the NLI task to contextualize word vectors. \citet{pan2018discourse} transfered  the  knowledge learned from the discourse marker prediction task
	to the NLI task to augment the semantic representation.

	\begin{figure*}[!t]
		\centering
		\includegraphics[width=0.9\textwidth]{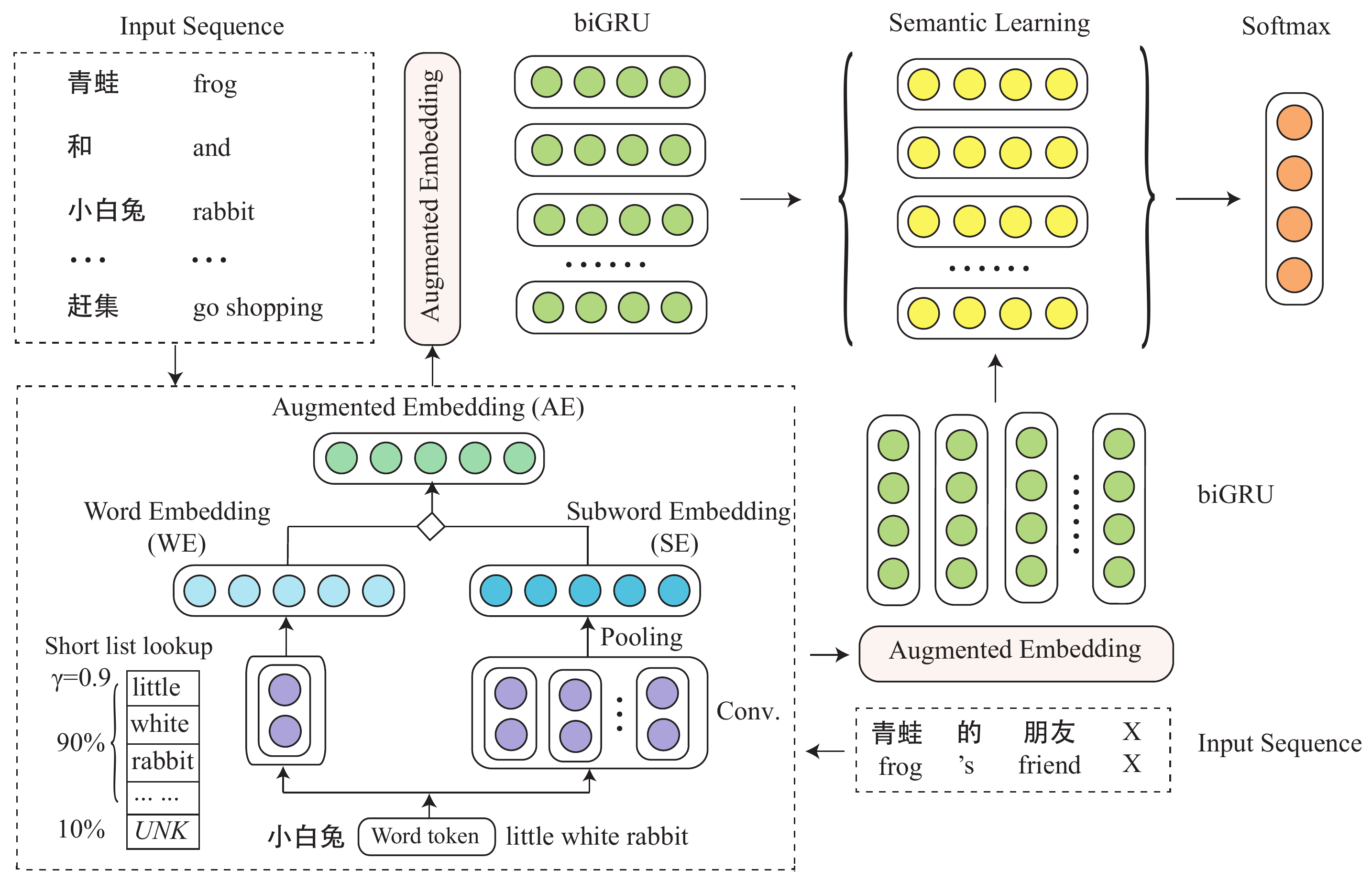}
	
		\caption{Architecture of the proposed Subword-augmented Embedding framework.}
		\label{fig:framework}
	\end{figure*}
	\section{Our Unified Representation Learning Framework}
	
	For generality, we consider an end-to-end model for either of text comprehension tasks. Fig. \ref{fig:framework} overviews the unified representation learning framework. The input tokens are segmented into subword units to further obtain the subword  embeddings, which are then fed to downstream models along with word embedding. For textual entailment, the two input sequences are premise and hypothesis and the output is the label. For reading comprehension, the two input sequences are document and question and the output is the answer.   
	
	We apply unsupervised subword segmentation to produce the subwords for each token in the input sequence. Our subwords are formed as character $n$-gram and do not cross word boundaries. After splitting each word $k$ into a subword sequence, an augmented embedding (AE) is formed to straightforwardly integrate word embedding $WE(k)$ and subword embedding $SE(k)$ for a given word $k$.
	\begin{align}	
	AE(k)= WE(k) \diamond  SE(k) 
	\end{align}
	where $\diamond$ denotes the integration strategy. In this work, we investigate concatenation (\emph{concat}), element-wise summation (\emph{sum}) and element-wise multiplication (\emph{mul}). 
	
	Suppose that word $k$ is formed with a sequence of subwords $[s_1, \dots, s_l]$ where $l$ is the number of subwords for word $k$. Then the subword-level representation of $k$ is given by the matrix $C^{k} \in R^{d\times l}$ where $d$ denotes the subword dimension.
	
	We employ a narrow convolution between $C^k$ and a filter $H \in R^{d \times w}$ of width $w$ to obtain a feature map $f^k \in R^{l-w+1}$. We take one filter operation for example, the $i$-th element of $f^k$ is given by
	\begin{align}
	f^{k}[i]=\tanh(\left \langle C^{k}[*, i:i+w-1], H \right \rangle + b)
	\end{align}  
	where $C^{k}[*, i:i+w-1]$ denotes the $i$-th to $(i+w-1)$-th column of $C^k$ and $\left \langle A,B \right \rangle$ = Tr($AB^T$) represents the Frobenius inner product. Then, a max pooling operation is adopted after the convolution and we fetch the feature representation corresponding to the filter $H$.
	\begin{align}
	y^{k}=\max_{i}f^{k}[i]
	\end{align}  
	
	Here we have described the process by which one feature is obtained from one filter matrix. For a total of $h$ filters, $[H_1, \dots, H_h]$, then $y^k = [y^{k}_1, \dots, y^{k}_h]$ is the distilled subword-level representation of word $k$. We then fed $y^k$ to a highway network \cite{Rupesh2015Training} to select features individually for each subword-derived word representation, and the final subword embedding (SE) is obtained by
	\begin{align}
	SE(k) &= t\odot g(W_{H}y^{k}+b_H) + (1-t)\odot y^{k}
	\end{align}
	where $g$ is a nonlinear function and $t=\sigma(W_{T}y^{k}+b_T)$ represents the transform gate. $W_{H}$, $W_{T}$, $b_H$ and $b_T$ are parameters.

	The downstream model is task-specific. In this work, we focus on the textual entailment and machine reading comprehension, which will be discussed latter.
	
	\subsection{Unsupervised Subword Segmentation}
	
	To segment subwords from word that is regarded as character sequence, we adopt and extend the generalized unsupervised segmentation framework proposed by \citet{zhao-ijcnlp2008}, which was originally designed only for Chinese word segmentation.
	
	The generalized framework can be divided into two collocative parts, \emph{goodness measurement} which evaluates how likely a subword is to be a `proper' one, and a \emph{segmentation} or \emph{decoding algorithm}. The framework generally works in two steps. First, a goodness score $g(w_{i})$ is computed for each $n$-gram $w_{i}$ (in this paper gram always refers to character) using the selected goodness measure to form a dictionary $W=\left \{ \left \{ w_{i}, g(w_{i}) \right \}_{i=1,\dots,n} \right \}$. Then segmentation or decoding method is applied to tokenize words into subwords based on the dictionary.

	\citet{zhao-ijcnlp2008} originally considered two decoding algorithms.
	
	\paragraph{Viterbi} This style of segmentation is to search for a segmentation with the largest goodness score sum for an input unsegmented sequence $T$ (to be either words or Chinese sentence).
	
	\paragraph{Maximal-Matching (MM)}
	This is a greedy algorithm with respect to a goodness score. It works on $T$ to output the best current subword
	$w^{*}$ repeatedly with $T =t^{*}$ for the next round as follows,
	\begin{align}
	\left \{ w^{*},t^{*} \right \}=\mathop{\arg\max}_{wt=T} g(w) 
	\end{align}
	with each $\left \{ w,g(w) \right \} \in W$.
	
	In this work, we additionally introduce the second segmentation algorithm.
	
	\paragraph{Byte Pair Encoding (BPE)}
	Byte Pair Encoding (BPE) \cite{Gage1994A} is a simple data compression technique that iteratively replaces the most frequent pair of bytes in a sequence by a single, unused byte. Different from the previous two algorithms that segment the input sequence into pieces in a top-down way, BPE segmentation actually merges a full single-character segmentation to a reasonable segmentation in a bottom-up way. We formulize the generalized BPE style segmentation in the following.
	
	At the very beginning, all the input sequences are tokenized into a sequence of single-character subwords, then we repeat,
	\begin{enumerate}
		\item Calculate the goodness scores of all bigrams under the current segmentation status of all sequences.
		\item Find the bigram with the highest goodness score and merge them in all the sequences. Note the segmentation status has been updated at this time.
		\item If the merging times does not reach the specified number, go back to 1, otherwise the algorithm ends.
	\end{enumerate}

	In our work, we investigate three types of goodness measures to evaluate subword likelihood, namely \emph{Frequency}, \emph{Accessor Variety} and \emph{Description Length Gain} \footnote{\citet{zhao-ijcnlp2008} considered four types of goodness measures but \emph{Branch Entropy} is excluded here due to its similar performance as \emph{Accessor Variety} according to their results}. 
	
	\paragraph*{Frequency (FRQ)}  FRQ is simply defined as the counting in the entire corpus for each $n$-gram being subword candidate. We take a logarithmic form as the goodness score,
	\begin{eqnarray}\label{eq-ijcnlp08-g-fsr}
	g_{_{FSR}}(w)={\rm log}(\hat{p}(w))
	\end{eqnarray}
	where $\hat{p}(w)$ is $w$'s frequency in the corpus.
	
	\paragraph*{Accessor Variety (AV)}
	AV is proposed by \citet{Feng2004Accessor} to measure how likely a subword is a true word. The AV of a subword $x_{i}x_{i+1} \ldots x_{j}$ (also denoted as $x_{i..j}$)  is defined as
	\begin{equation}\label{eq:av}
	AV(x_{i..j}) = \min \{L_{av}(x_{i..j}),R_{av}(x_{i..j})\}
	\end{equation}	
	where the left and right accessor variety $L_{av}(x_{i..j})$
	and $R_{av}(x_{i..j})$ are the number of distinct predecessor and successor characters, respectively. The same as FRQ, the goodness score is taken in logarithmic form, $g_{AV}(w) = {\rm log} AV (w)$.
	
	\paragraph*{Description Length Gain (DLG)}
	\citet{Wilks1999Unsupervised} proposed this goodness measure for compression-based segmentation. The DLG replaces all occurrences of $x_{i..j}$ from a corpus $X=x_1x_2...x_n$
	as a subword and is computed by
	\begin{equation}\label{eq:dlg}
	DLG(x_{i..j}) = L(X)-L(X[r \to x_{i..j}]  \oplus x_{i..j})
	\end{equation}
	where $X[r \to x_{i..j}]$ represents the resultant corpus by replacing all items of $x_{i..j}$ with a new symbol $r$ throughout $X$ and $\oplus$ denotes the concatenation.
	$L(\cdot)$ is the empirical description length of a corpus in bits that can be estimated by the Shannon-Fano code or Huffman code, following classic information theory \cite{BLTJ:BLTJ1338}.
	\begin{equation}\label{eq:shannon}
	L(X) \doteq -|X|\sum_{x \in V}\hat{p}(x) \log_{2}\hat{p}(x)
	\end{equation}
	where $|\cdot|$ denotes the string length, $V$ is the vocabulary of $X$ and $\hat{p}(x)$ is $x$'s frequency in $X$. The goodness score is given by $g_{DLG}(w) =DLG(w)$.
	
	It is easy to find that BPE style segmentation with FRQ goodness measures (denoted as BPE-FRQ) could be identical to the BPE subword encoding in \cite{Sennrich2015Neural} in neural machine translation which is originally motivated for word representation for infrequent (rare or OOV) word representation in neural machine translation. Instead, we aim to refine the word representations by using subwords, for both frequent and infrequent words, which is more generally motivated. To this end, we adaptively tokenize words in multi-granularity.

	\section{Experiments}
	In this section, we evaluate the performance of subword-augmented embedding on two kinds of challenging text understanding tasks, \emph{textual entailment} and \emph{reading comprehension}. Both of the concerned tasks are quite challenging, let alone the latest performance improvement has been already very marginal. However, we present a new solution in a new direction instead of heuristically stacking attention mechanisms. Namely, we show that subword embedding could be potential to give further advances due to its meaningful linguistic augments, which has not been studied yet for the concerned tasks. Our evaluation aims to answer the following empirical questions:

	\begin{enumerate}
		\item Can subword-augmented embedding enhance the concerned tasks?
		\item  Can using subword-augmented embedding be generally helpful for different languages?
		\item  Can subword embedding help effectively model OOV or rare words?
		\item Which is the best unsupervised subword segmentation method  for text understanding?
		\item Which is the best strategy to integrate word and subword embedding?
	\end{enumerate}
	
	The default subword vocabulary size is set 10$k$ for textual entailment task and 1$k$ for the two reading comprehension tasks. The default integration strategy is \emph{concatenation} for the following experiments. The above choices are based on the model performance on the development set and the detailed analysis will be given in Section 6. Word embeddings are 200$d$ and pre-trained by word2vec \cite{mikolov:2013} toolkit on \emph{Wikipedia} corpus\footnote{\url{https://dumps.wikimedia.org/} }. Both character and subword embeddings are also 200$d$ and randomly initialized with the uniform distribution in the interval [-0:05; 0:05]. Note that character could be regarded as the minimal case of subwords so we separately depict them in our experiments for better comparison and convenient demonstration. 
	
In our preliminary experiments, we thoroughly explore all nine subword segmentation methods by considering there are three segmentation algorithms and three goodness measures. We find that all Viterbi based segmentation fails to show satisfactory performance, and we only report three best performing segmentation-goodness collocations for each task.  Our baseline models are selected due to their simplicity and state-of-the-art performance in each task. We are interested in a subword-based framework that performs robustly across a diverse set of tasks. To this end, we follow the same hyper-parameters or each baseline model as the original settings from their corresponding literatures \cite{Chen2017Enhanced,Dhingra2017Gated,Seo2016Bidirectional} except those specified (e.g. subword dimension, integration strategy). Since ensemble systems and pre-training enhanced methods are commonly integrated with multiple heterogeneous models and resources and thus not completely comparable, we only focus on the evaluations on single models.
	
	\subsection{Textual Entailment}
	
	\begin{table}
		\centering
		\caption{\label{tab:snli} Accuracy on SNLI dataset. SOTA is short for state-of-the-art. }
		{
			\begin{tabular}{l c c}
				\toprule
				Model& Dev & Test\\
				\hline
				Baseline (Word + Char) &  88.39  & 87.61\\
				\hline
				Word + Viterbi-AV & 88.35  & 87.70 \\	
				Word + Viterbi-FRQ & 88.15 & 87.46 \\
				Word + Viterbi-DLG & 88.31 & 87.53 \\
				\hline
				Word + MM-AV & 88.58  & 88.16 \\	
				Word + MM-FRQ & 88.45 & 88.05 \\
				Word + MM-DLG & 88.61 & 88.28 \\
				\hline
				Word + BPE-AV &  88.42  &  88.11\\	
				Word + BPE-FRQ &  88.56 &  88.36\\
				Word + BPE-DLG & \textbf{88.68}   &  \textbf{88.56}\\
				\hline
				SOTA \cite{Kim2018Semantic}  &  /  &  88.9\\
				\bottomrule	
			\end{tabular}
		}
	\end{table}

	\begin{table*}
		\centering
		\caption{\label{tab:dataset} Data statistics of CMRC-2017, PD and CFT.}
		{
			\begin{tabular}{cccccccc}
				\hline
				\hline
				& \multicolumn{3}{c}{CMRC-2017} & \multicolumn{3}{c}{PD}  & CFT  \\
				&Train & Valid & Test & Train & Valid & Test & human  \\
				\hline
				\# Query & 354,295 & 2,000 & 3,000 &870,710 & 3,000 & 3,000  & 1,953\\
				Max \# words in docs  & 486 & 481  & 484 & 618 & 536 & 634 &  414\\
				Max \# words in query & 184 & 72  & 106 & 502 & 153 & 265 &   92\\
				Avg \# words in docs & 324 & 321  & 307 & 379 & 425 & 410 &   153\\
				Avg \# words in query & 27 & 19 & 23 & 38 & 38 & 41 & 20\\
				\# Vocabulary  & 94,352 & 21,821 & 38,704 & 248,160 & 536 & 634 & 414\\
				\hline
				\hline
			\end{tabular}
		}
	\end{table*}

	Textual entailment is the task of determining whether a \emph{hypothesis} is \emph{entailment, contradiction} and \emph{neutral}, given a \emph{premise}. The Stanford Natural Language Inference (SNLI) corpus \cite{Bowman2015A} provides approximately 570$k$ hypothesis/premise pairs. 
	
	Our baseline model is Enhanced Sequential Inference Model (ESIM) \cite{Chen2017Enhanced} which employs a biLSTM to encode the premise and hypothesis, followed by an attention layer, a local inference layer, an inference composition layer. To keep the model simplicity and concentrate on the performance of subword units, we do not integrate extra syntactic parsing features or increase the dimension of word embeddings. However, with the subword augmentation, our simple sequential encoding model yields substantial gains and achieves competitive performance with more complex state-of-the-art models\footnote{We only compare with currently published work from SNLI Leaderboard: \url{https://nlp.stanford.edu/projects/snli/}}.
	
	The dimensions for all the LSTM and fully connection layers were 300. We set the dropout rate to 0.5 for each LSTM layer and the fully connected layers. All feed forward layers used ReLU activations. Parameters were optimized using Adam  \cite{kingma2014adam} with gradient norms clipped at 5.0. The initial learning rate was 0.001, which was halved every epoch after the second epoch. The batch size was 32.
	
	Results in Table \ref{tab:snli} show that, subword-augmented embedding can boost our baseline (Word + Char) by +0.95\% on the test set. Among the subword algorithms, BPE-DLG performs the best whose key difference with other approaches is that BPE-DLG gives finer-grained bi-grams like \{\emph{ri, ch, ne, ss}\} which could be potentially important for short text modeling with small word vocabulary like textual entailment task.

	\subsection{Reading Comprehension}

	To investigate the effectiveness of the subword-augmented embedding in conjunction with more complex models, we conduct experiments on machine reading comprehension tasks. The reading comprehension task can be described as a triple $<D, Q, A>$, where $D$ is a document (context), $Q$ is a query over the contents of $D$, in which a word or span is the right answer $A$. This task can be divided into cloze-style and query-style. The former has restrictions that each answer should be a single word and should appear in the document and the original sentence removing the answer part is taken as the query. For the query-style, the query is formed as questions in natural language whose answer is a span of texts. To test the subword-augmented embedding in multi-lingual case, we select three Chinese datasets, \emph{Chinese Machine Reading Comprehension (CMRC-2017)} \cite{Cui2017Dataset}, \emph{People's Daily (PD)} \cite{Cui2016Consensus}, \emph{Children Fairy Tales (CFT)} \cite{Cui2016Consensus} and two English ones, \emph{Children's Book Test (CBT)} \cite{hill2015goldilocks}, \emph{the Stanford Question Answering Dataset (SQuAD)} \cite{Rajpurkar2016SQuAD} in which the first four sets are cloze-style and the last one is query-style.

	\subsubsection{Cloze-style}
		
	\begin{table}
		\centering
		\caption{\label{tab:cmrc} Accuracy on CMRC-2017 dataset.}
		\begin{tabular}{l c c}
			\toprule
			Model
			& Dev & Test\\
			\hline
			Baseline (Word + Char) & 76.15 & 77.73\\
			\hline
			Word + MM-AV & 77.80 & 77.80\\	
			Word + MM-DLG & 77.30 & 77.17\\
			Word + BPE-FRQ & \textbf{78.95}  & \textbf{78.80}\\
			\hline
			SOTA \cite{Cui2017Attention}  & 77.20  & 78.63\\
			\bottomrule	
		\end{tabular}
	\end{table}

	To verify the effectiveness of our proposed model for Chinese, we conduct multiple experiments on three Chinese Machine Reading Comprehension datasets, namely CMRC-2017, PD and CFT \footnote{Note that the test set of CMRC-2017 and human evaluation test set (Test-human) of CFT are harder for the machine to answer because the questions are further processed manually and may not be accordance with the pattern of automatic questions.}. Table \ref{tab:dataset} gives data statistics. Different from the current cloze-style datasets for English reading comprehension, such as CBT, Daily Mail and CNN \cite{hermann2015teaching}, the three Chinese datasets do not provide candidate answers. Thus, the model has to find the correct answer from the entire document.
	
	Our baseline model is the Gated-Attention (GA) Reader \cite{Dhingra2017Gated} which integrates a multi-hop architecture with a gated attention mechanism between the intermediate states of document and query. We used stochastic gradient descent with ADAM updates for optimization. The batch size was 32 and the initial learning rate was 0.001 which was halved every epoch after the second epoch. We also used gradient clipping with a threshold of 10 to stabilize GRU training (Pascanu \emph{et al.}, 2013). We used three attention layers. The GRU hidden units for both the word and subword representation were 128. We applied dropout between layers with a dropout rate of 0.5.

	\paragraph{CMRC-2017} Table \ref{tab:cmrc} gives our results on CMRC-2017 dataset \footnote{CMRC-2017 Leaderboard: \url{http://www.hfl-tek.com/cmrc2017/leaderboard/}.}, which shows that our Word + BPE-FRQ model outperforms all other models on the test set, even the state-of-the-art AoA Reader \cite{Cui2017Attention}. With the help of the proposed method, the GA Reader could yield a new state-of-the-art performance over the dataset. Different from the above textual entailment task, the best subword segmentation tends to be BPE-FRQ instead of BPE-DLG. The divergence indicates that for a task like reading comprehension involving long paragraphs with a huge vocabulary \footnote{The word vocabulary sizes of SNLI and CMRC-2017 are 30k and 90k respectively.}, high frequency words weigh more. In fact, as DLG measures word through more type statistics than the direct frequency weighting, it can be seriously biased by a lot of noise in the vocabulary. Using frequency instead of DLG can let the segmentation resist the noise by keeping concerns over those high frequency (also usually regular) words. Since we found the stable performance gain in all our preliminary experiments, we focus on BPE-FRQ in later similar cloze-style evaluation and comparison.

	\begin{table}
		\centering 
		\caption{\label{tab:pdcftresult} Accuracy on PD and CFT datasets. Results of AS Reader and CAS Reader are from \cite{Cui2016Consensus}. The result for GA Reader is based on our implementation. Previous state-of-the-art model is marked by $\dagger$.}
		{
			\begin{tabular}{l|c|c|c}
				\hline
				\hline
				\multirow{2}{*}{Model}  & \multicolumn{2}{c}{PD}  & CFT  \\
				& Valid & Test &Test-human  \\
				\hline
				AS Reader &  64.1 & 67.2 & 33.1  \\
				CAS Reader$\dagger$&  65.2 & 68.1 & 35.0  \\
				GA Reader &  67.2 & 69.0 & 36.9  \\
				\hline
				Word + BPE-FRQ & \textbf{72.8} & \textbf{75.1} & \textbf{43.8}    \\
				\hline
				\hline
			\end{tabular}
		}
	\end{table}

	\paragraph{PD \& CFT} Since there is no training set for CFT dataset, our model is instead trained on PD training set. Note that CFT test set is processed by human evaluation, and may not be accordance with the pattern of PD training dataset. The results on PD and CFT datasets are listed in Table \ref{tab:pdcftresult}, which shows our Word + BPE-FRQ significantly outperforms the CAS Reader in all types of testing, with improvements of 7.0\% on PD and 8.8\% on CFT test sets, respectively. Considering that the domain and topic of PD and CFT datasets are quite different, the results indicate the effectiveness of our model for out-of-domain learning.

	\begin{table}
		\centering 
		\caption{\label{tab:cbt} Accuracy on CBT dataset. Results except ours are from previously published works \cite{Dhingra2017Gated,Cui2016Consensus,Yang2016Words}. Previous state-of-the-art model is marked by $\dagger$.}
		{
			\begin{tabular}{l|c|c|c|c}
				\hline
				\hline
				\multirow{2}{*}{Model}  & \multicolumn{2}{c}{CBT-NE}  & \multicolumn{2}{c}{CBT-CN} \\
				& Valid & Test   & Valid & Test  \\
				\hline
				Human  &- & 81.6 & - & 81.6 \\
				\hline
				LSTMs  & 51.2 & 41.8 & 62.6 & 56.0 \\ 
				MemNets  & 70.4 & 66.6 & 64.2 & 63.0 \\
				AS Reader & 73.8 & 68.6 & 68.8 & 63.4 \\
				Iterative Attentive Reader & 75.2 & 68.2 & 72.1 & 69.2 \\
				EpiReader  & 75.3 & 69.7 & 71.5 & 67.4 \\
				AoA Reader  & 77.8 & 72.0 & 72.2 & 69.4 \\
				NSE & 78.2 & 73.2 & 74.3 & 71.9 \\
				FG Reader$\dagger$  & \textbf{79.1} & \textbf{75.0} & \textbf{75.3}&  \textbf{72.0} \\
				GA Reader  &76.8 & 72.5  & 73.1 & 69.6 \\
				\hline
				Word + BPE-FRQ& 78.5 & 74.9 & 75.0 & 71.6\\
				\hline
				\hline
			\end{tabular}
		}
		
	\end{table}	
	\paragraph{CBT} To verify if our method can work for more than Chinese, we also evaluate the proposed method on English benchmark, CBT, whose documents consist of 20 contiguous sentences from the body of a popular children's book, and queries are formed by deleting a token from the 21st sentence. We only focus on its subsets where the answer is either a common noun (CN) or NE, so that our task here is more challenging as the answer is likely to be rare words. For a fair comparison, we simply set the same parameters as before. We evaluate all the models in terms of accuracy, which is the standard evaluation metric for this task. 
	
	Table \ref{tab:cbt} shows the results for CBT. We observe that our model outperforms most of the previously published works, with 2.4 \% gains on the CBT-NE test set compared with GA Reader which adopts word and character embedding concatenation. Our Word + BPE-FRQ also achieves comparable performance with FG Reader who adopts neural gates to combine word-level and character-level representations with assistance of extra features including NE, POS and word frequency while our model is much simpler and faster. This comparison shows that our Word + BPE-FRQ is not restricted to Chinese reading comprehension, but also effective for other languages.
	
	\subsubsection{Query-style}
	\begin{table}
		\centering
		\caption{\label{tab:squad} Exact Match (EM) and F1 scores on SQuAD dev set. BiDAF$_\alpha$ denotes \emph{BiDAF + Self-Attention} and BiDAF$_\beta$ denotes \emph{BiDAF + Self-Attention + ELMO}.}
		{
			\begin{tabular}{l l c c}
				\toprule
				\multicolumn{2}{c}{Model} & EM & F1  \\
					\hline
				\multirow{4}{*}{BiDAF$_\alpha$}& Word + Char & 71.22 & 80.42     \\
				& Word + MM-AV  & 72.46 & 81.28     \\
				& Word + MM-DLG &  72.21 & 81.03       \\
				& Word + BPE-FRQ & \textbf{72.79}  & \textbf{81.78} \\
				\hline
				\multirow{4}{*}{BiDAF$_\beta$}& Word + Char & 77.43 & 85.03      \\
				& Word + MM-AV    & 77.49  & 85.23 \\
				&Word + MM-DLG & 77.46 & 85.22     \\
				& Word + BPE-FRQ & \textbf{77.84} & \textbf{85.48}  \\
				\hline
				\multirow{4}{*}{BiDAF} & Word + Char & 68.23 & 77.95      \\
				& Word + MM-AV    & 68.86  & 78.44 \\
				&Word + MM-DLG & 68.82 & 78.40     \\
				& Word + BPE-FRQ & \textbf{69.35} & \textbf{78.97}  \\
				\bottomrule
			\end{tabular}
		}
	\end{table}
	The Stanford Question Answering Dataset (SQuAD) \cite{Rajpurkar2016SQuAD} contains 100$k$+ crowd sourced question-answer pairs where the answer is a span in a given Wikipedia paragraph. Our basic model is  Bidirectional Attention Flow \cite{Seo2016Bidirectional} and we improve it by adding  a self-attention layer \cite{Wang2017Gated} and ELMO \cite{Peters2018ELMO}, similar to \cite{clark2018simple}, to see whether subword could still improve more complex models. The augmented embeddings of document and query are  passed through a bi-directional GRU which share parameters, and then fed to the BiDAF model. Then, we obtain the context vectors and pass them through a linear layer with ReLU activations, followed by a self-attention layer against the context itself. Finally, the results are fed through linear layers to predict the start and end token of the answer. For the hyper-parameters, the dropout rates for the GRUs and linear layers are 0.2. The dimensions for GRU and linear layers are 90 and 180, respectively. We optimize the model using ADAM. The batch size is 32. Table \ref{tab:squad} shows the results on the dev set \footnote{Since the test set is not released, we train our models on training set and evaluate them on dev set.}. We can see that for all the models, subword embeddings boost the performance significantly. Even for BiDAF$_\alpha$ and BiDAF$_\beta$, BPE-FRQ could also yield substantial performance gains (+1.57\%EM, 1.36\%F1 and +0.41\%EM, 0.45\%F1 respectively).

	\begin{table}
		\centering
		\caption{\label{tab:abs} Embedding combinations on CMRC-2017.}
		\begin{tabular}{l c c}
			\toprule
			Model
			& Dev & Test \\
			\hline
			Word Only & 74.90 & 75.80\\
			Char Only & 71.25 & 72.53\\
			BPE-FRQ Only & 74.75 & 75.77\\
			\hline
			Word + Char & 76.15 & 77.73 \\		
			Word + BPE-FRQ & 78.95  & 78.80 \\		
			Word + Char + BPE-FRQ & \textbf{79.05} & \textbf{78.83}\\
			\bottomrule	
		\end{tabular}
	\end{table}

	\begin{figure*}
		\subfigure[SNLI]{
			\begin{minipage}[b]{0.3\textwidth}
				\includegraphics[width=1\textwidth]{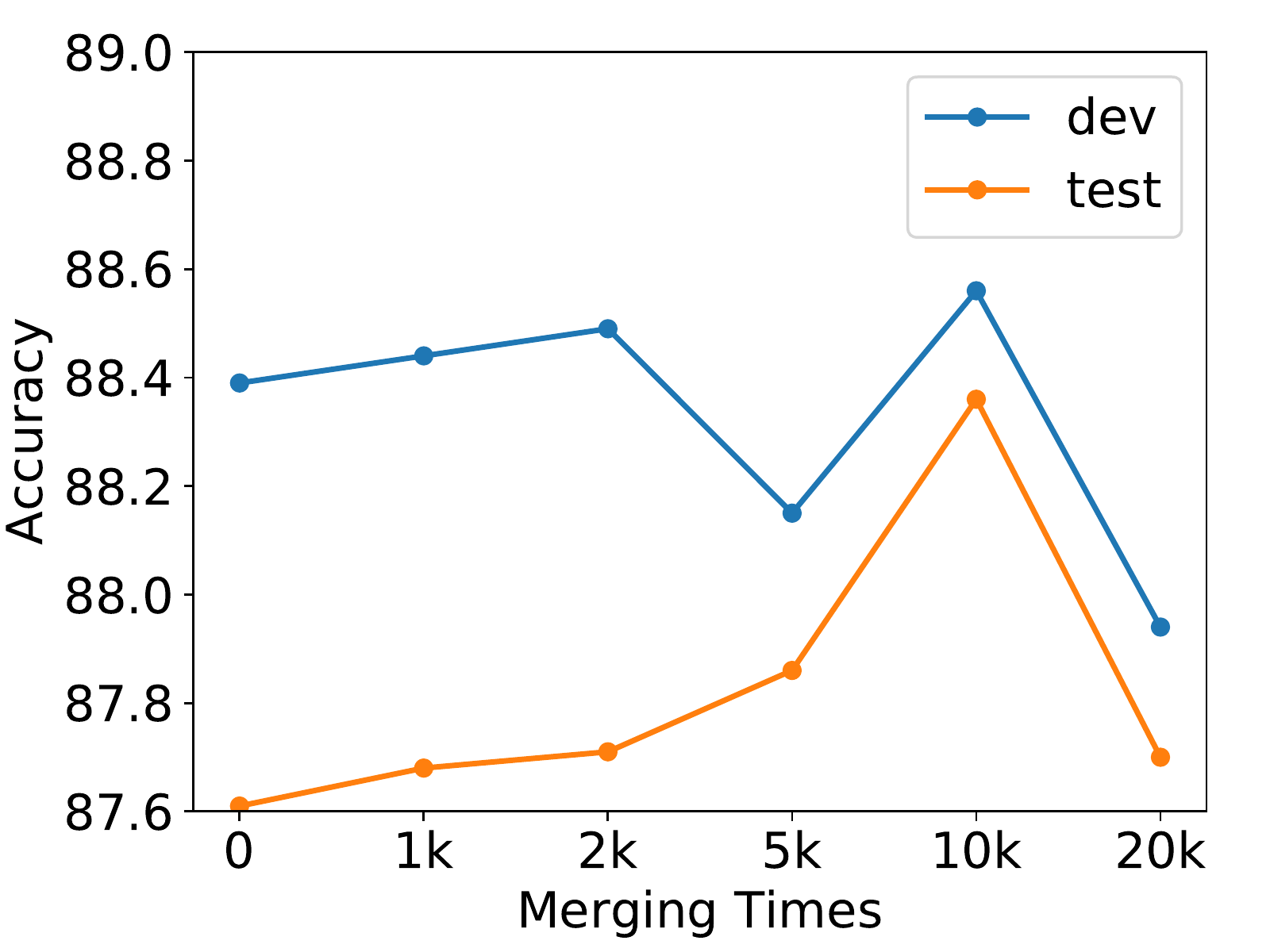}
			\end{minipage}
		}
		\subfigure[CMRC-2017]{
			\begin{minipage}[b]{0.305\textwidth}
				\includegraphics[width=1\textwidth]{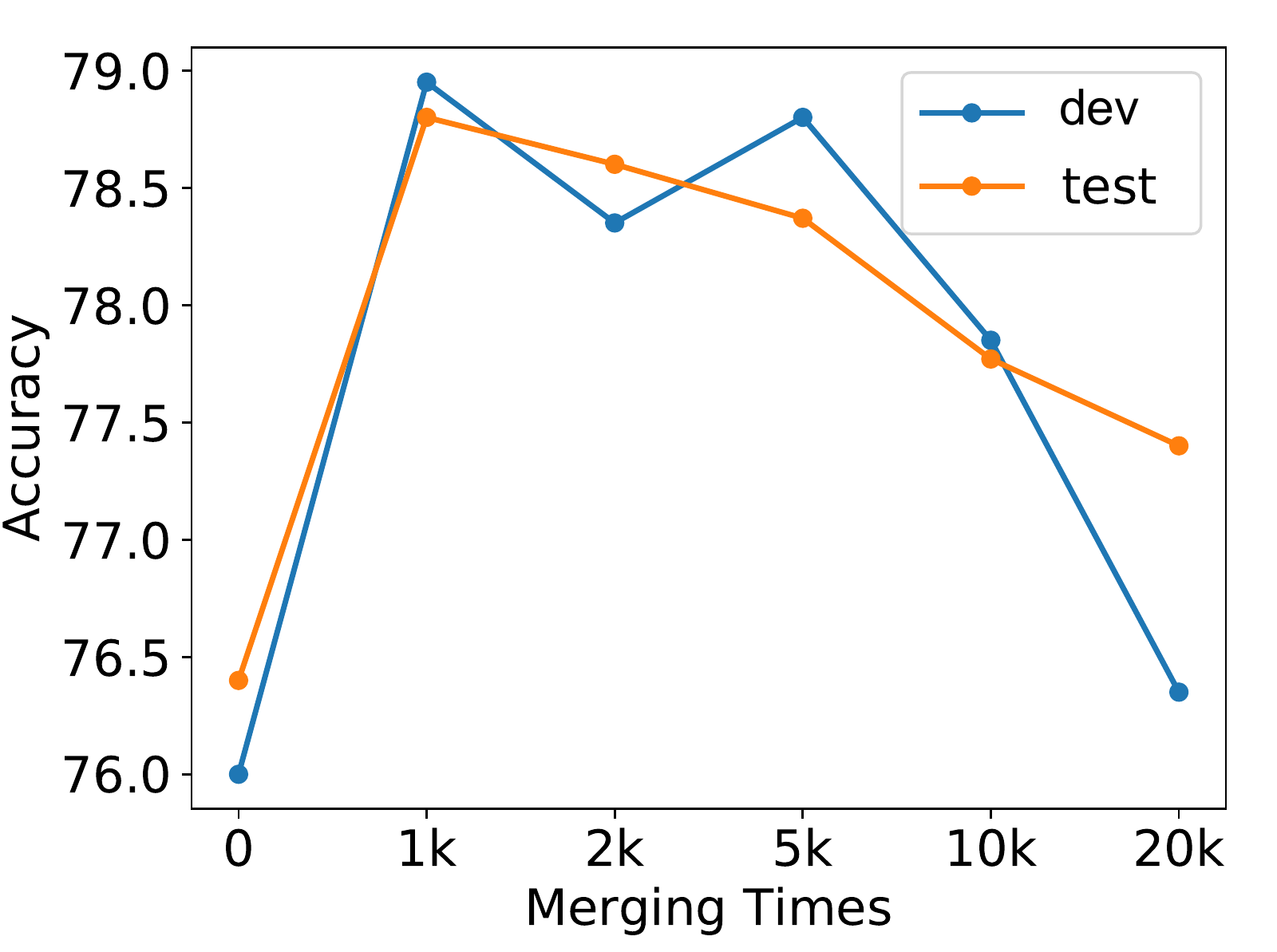}
			\end{minipage}
		}
		\subfigure[SQuAD]{
			\begin{minipage}[b]{0.3\textwidth}
				\includegraphics[width=1\textwidth]{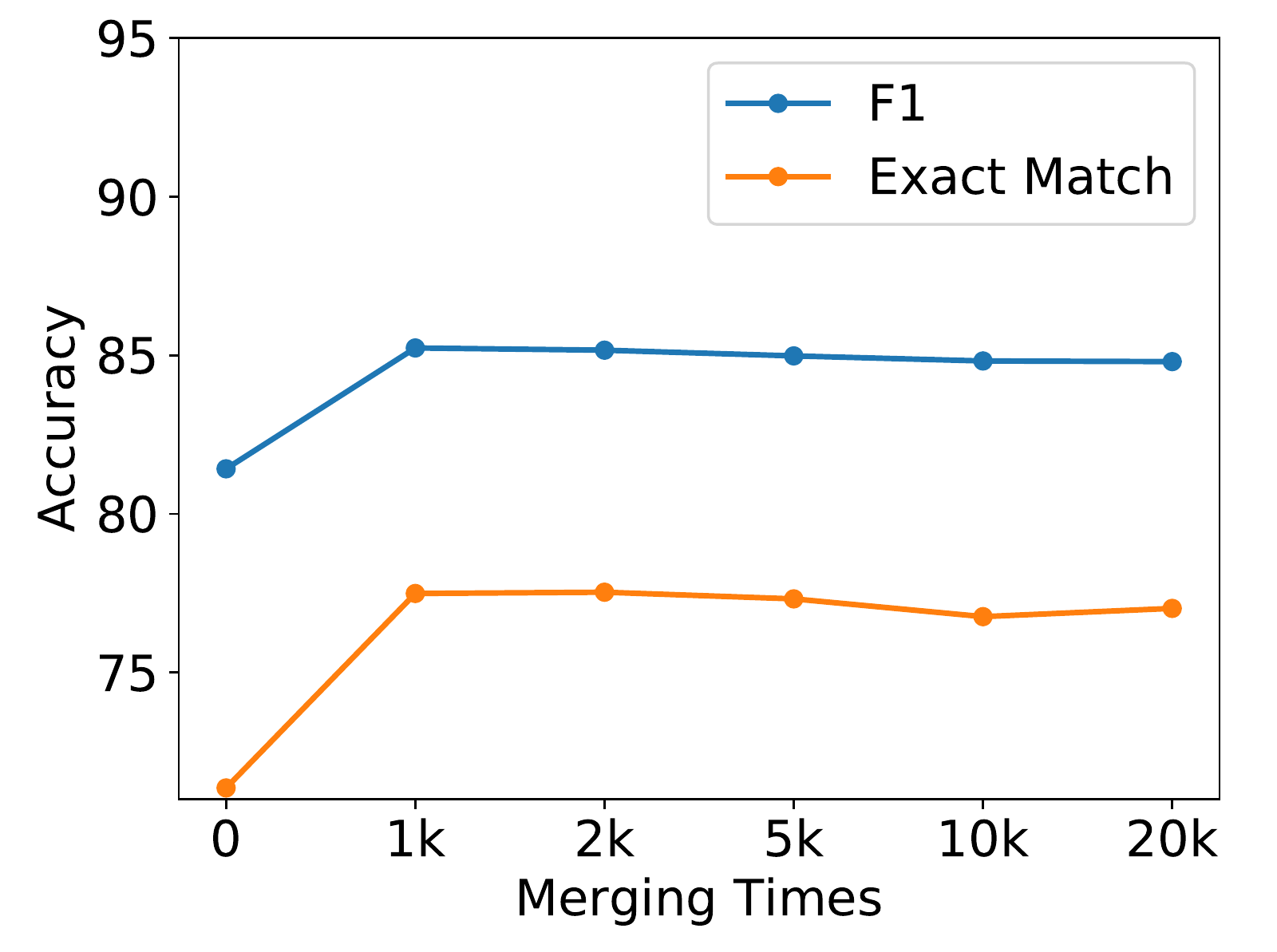}
			\end{minipage}
		}
		\caption{Case study of the subword vocabulary size of BPE-FRQ.} \label{fig:merge}	
	\end{figure*} 
	\section{Analysis}
	The experimental results have shown that the subword-augmented embedding can essentially improve baselines, from the simple to the complicated, among multiple tasks with different languages. Though the performance of BPE-FRQ tends to be the most stable overall, the best practice for subword embedding might be task-specific. This also discloses that there exists potential for a more effective goodness measure or segmentation algorithm to polish up the subword representations.
	
	\subsection{Using Diverse Embedding Together}
	To see if we can receive further performance improvement when using different embedding together, we compare the following embeddings: Word Only, Char Only, BPE-FRQ only and Word + Char, Word + BPE-FRQ and Word + Char + BPE-FRQ. Table \ref{tab:abs} shows the result. For each type of embedding alone, word embedding and BPE-FRQ subword embedding turn out to be comparable. BPE-FRQ performs much better than char embedding, which again confirms that subwords are more representative as minimal natural linguistic units than single characters. Any embedding combination could improve the performance as the distributed representations can be beneficial from different perspectives through diverse granularity. However, using all the three types of embeddings only shows marginal improvement. This might indicate that increasing embedding features or dimension might not bring much gains and seeking natural and meaningful linguistic units for representation is rather significant.

	\subsection{Subword Vocabulary Size}
	The segmentation granularity is highly related to the subword vocabulary size. For BPE style segmentation, the resulting subword vocabulary size is equal to the merging times plus the number of single-character types. To have an insight of the influence, we adopt merge times of BPE-FRQ from 0 to 20$k$, and conduct quantitative study on SNLI, CMRC-2017 and SQuAD for BPE-FRQ segmentation. Fig. \ref{fig:merge} shows the results. We observe that with 1$k$ merge times, the models could obtain the best performance on CMRC-2017 and SQuAD though these two tasks are of different languages while 10k shows to be more suitable for SNLI. The results also indicate that for a task like reading comprehension the subwords, being a highly flexible grained representation between character and word, tends to be more like characters instead of words. However, when the subwords completely fall into characters, the model performs the worst. This indicates that the balance between word and character is quite critical and an appropriate grain of character-word segmentation could essentially improve the word representation. 

	\begin{table}
		\centering
		\caption{\label{tab:oper} Different merging functions with word embeddings on SNLI and CMRC-2017.}
		\begin{tabular}{l c c c}
			\toprule
			Model & Strategy & Dev & Test\\
			\hline
			& concat  & \textbf{88.68}   &  \textbf{88.56} \\
			SNLI & sum   & 88.30   & 87.14  \\
			& mul  &88.47 & 87.77 \\
			\hline
			& concat   &  77.45 & 77.47   \\
			CMRC & sum   & 75.95  & 76.43  \\
			& mul  & \textbf{78.95} &  \textbf{78.80} \\
			\bottomrule
		\end{tabular}
		
	\end{table}

	\begin{figure}
		\centering
		\includegraphics[width=0.38\textwidth]{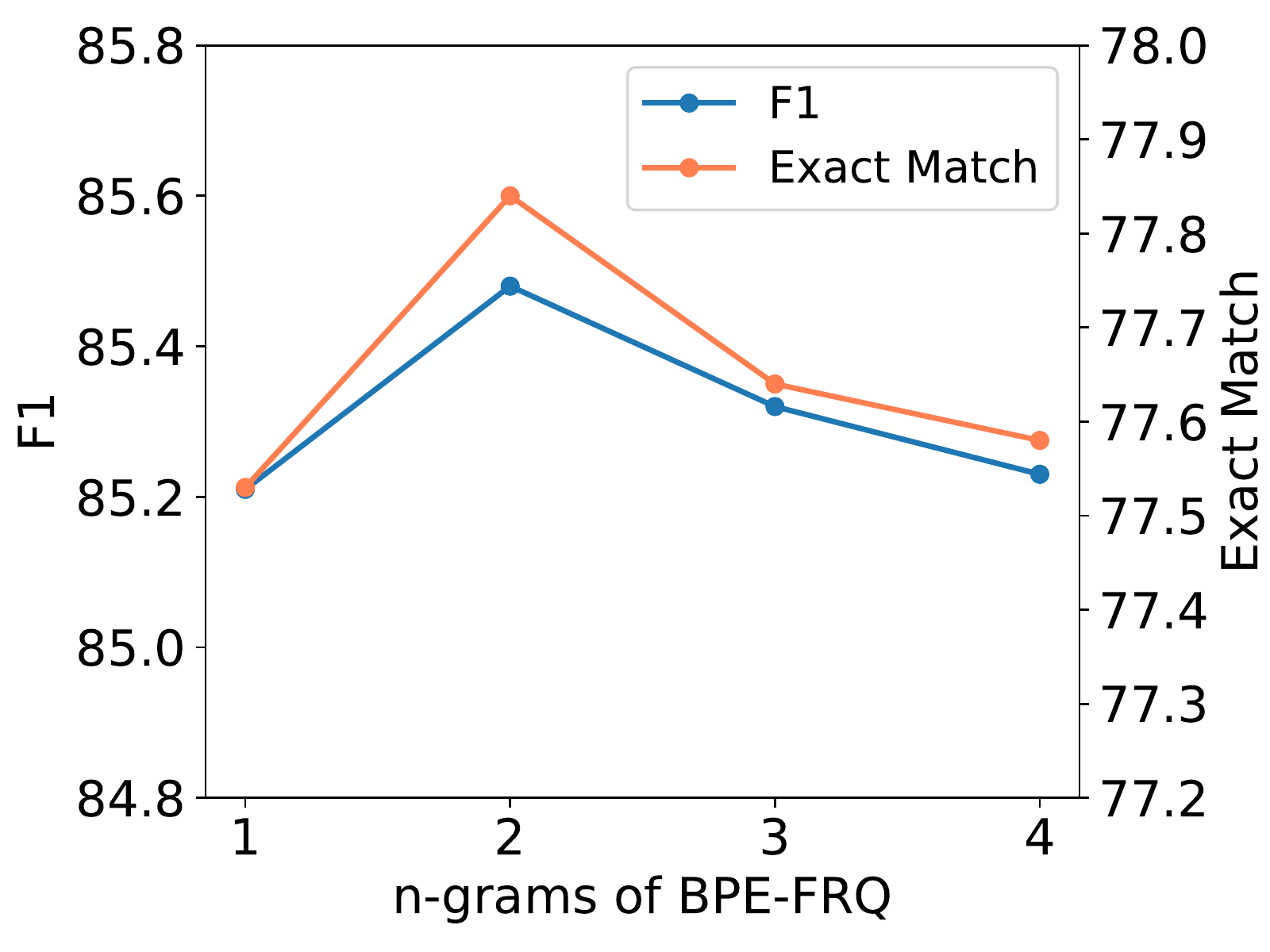}
		\caption{\label{fig:ngram}Results of $n$-gram of BPE-FRQ on SQuAD dataset.}
	\end{figure}

	\begin{figure*}\centering
		\caption{Pair-wise attention visualization. } \label{fig:attention}	
		\subfigure[Embedding of document and query]{\centering
			\begin{minipage}[b]{0.38\textwidth}
				\includegraphics[width=1\textwidth]{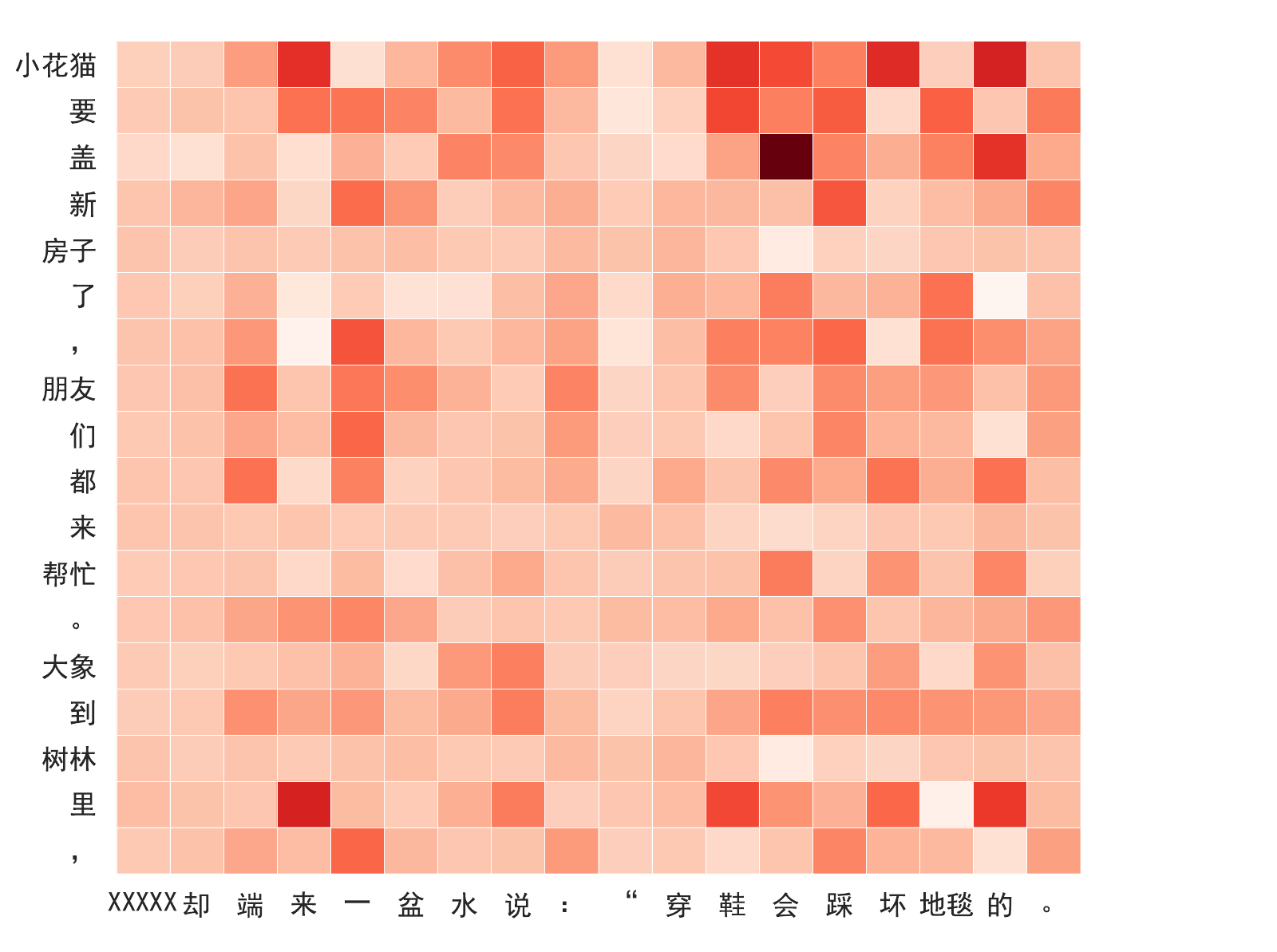}
			\end{minipage}
		}
		\subfigure[Final document and query representation]{\centering
			\begin{minipage}[b]{0.38\textwidth}
				\includegraphics[width=1\textwidth]{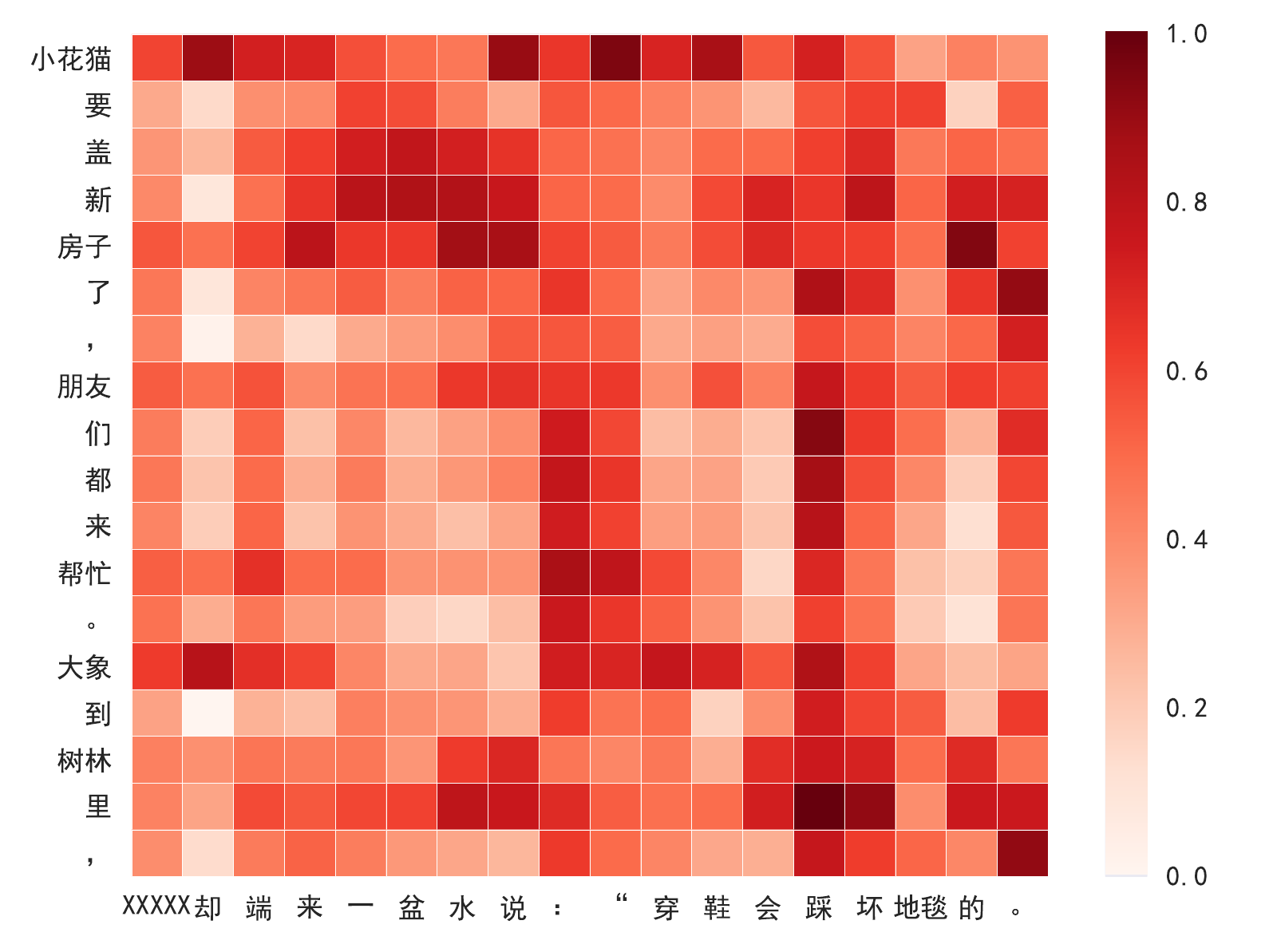}
			\end{minipage}
		}
		\\\begin{flushleft}
			\scriptsize{\emph{Doc (extract): The cat was going to build a new house. His friends came to help. The elephant went into the woods for logs. The goat and dog cut the logs into planks. Soon afterwards the bear constructs a beautiful house. The cat said happily, ``After decorating my house, I'll invite everybody to have a party in it." A few days later, friends came to the party happily. Upon entering the door, the cat fetched a small basin of water and said, ``Your shoes will trample the carpet. Please take off your shoes and wash your feet, or leave." The elephant and bear looked at their own feet and the small basin, felt upset, saying, ``Forget it, we'll never go in." Since then, no animal played with him any more. The house was his last friend.
					\\Query: \underline{\hbox to 8mm{}}  fetched a small basin of water and said, ``Your shoes will trample the carpet.Please take off your shoes and wash your feet, or leave."} }\end{flushleft}
	\end{figure*}

	\subsection{Subword and Word Embedding Integration Strategies}
	We investigate the combination of subword-augmented embedding with word embedding. Table \ref{tab:oper} shows the comparisons based on our best models of SNLI and CMRC-2017, BPE-DLG and BPE-FRQ, respectively. The models with \emph{concat}  and \emph{mul} significantly outperform the model with \emph{sum}. This reveals that \emph{concat} and \emph{mul} operations might be more informative than \emph{sum}  and the best practice for the choice would be task-specific. Though \emph{concat} operation may result in high dimensions, it could keep more information for downstream models to select from. The superiority of \emph{mul} might be due to element-wise product being capable of modeling the interactions and eliminating distribution differences between word and subword embedding which is intuitively similar to endowing subword-aware \emph{attention} over the word embedding. In contrast, \emph{sum} is too simple to prevent from detailed information loss.

	\subsection{Effect of the $n$-grams}
	The goodness measures commonly build the subword vocabulary based on neighbored character relationship inside words. This is reasonable for Chinese where words are commonly formed by two characters which is also the original motivation for Chinese word segmentation. However, we wonder whether it would be better to use longer $n$-gram connections. We expand the \emph{$n$-grams} of BPE-FRQ from 1 to 4. Fig. \ref{fig:ngram} shows the quantitative study results. We observe the $n$-grams of BPE-FRQ segmentation might slightly influence the result where 2 or 3 tends to be better choice.

	\subsection{Visualization} 
	To analyze the learning process of our models, we draw the attention distributions at intermediate layers based on an example from CMRC-2017 dataset. Fig. \ref{fig:attention} shows the result of model with BPE-FRQ. We observe that the right answer (\emph{The cat}) could obtain a high weight after the pair-wise matching of document and query. After attention learning, the key evidence of the answer would be collected and irrelevant parts would be ignored. This shows that our subword-augmented embedding is effective at selecting the vital points at the fundamental embedding layer, guiding the attention layers to collect more relevant pieces.

	\subsection{Subword Observation}
	In text understanding tasks, if the ground-truth answer is OOV word or contains OOV word(s), the performance of deep neural networks would severely drop due to the incomplete representation, especially for a task like cloze-style reading comprehension where the answer is only one word or phrase. To get an intuitive observation for the task, we collect all the 118 questions whose answers are OOV words (with their corresponding documents, denoted as \emph{OOV questions}) from CMRC-2017 test set, and use our model to answer these questions. We observe only 2.54\% could be correctly answered by the best Word + Char embedding based model. With BPE-FRQ subword embedding, 12.71\% of these OOV questions could be correctly solved. This shows that the subword representations could be essentially useful for modeling rare and unseen words. In fact, the meaning of complex words like \emph{indispensability} could be accurately refined by segmented subwords as shown in Table \ref{tab:example_bck}. This also shows subwords could help the models to use morphological clues to form robust word representations which is especially potential to obtain fine-grained representation for low-resource languages.

	\begin{CJK*}{UTF8}{gkai}
		\begin{table}[h]
			\centering
			\caption{\label{tab:example_bck} Examples of BPE-FRQ subwords.}
			{
				\begin{tabular}{|l|l|}
					\hline
					Word & Subword \\
					\hline
					indispensability & in disp ens ability \\
					intercontinentalexchange &	inter contin ent al ex change \\
					playgrounds &	play ground s\\
					大花猫	& 大 $\;$花猫 \\
					一步一个脚印	& 一步$\ $一个$\ $脚印 \\
					\hline
				\end{tabular}
			}
			
		\end{table}
	\end{CJK*}

	\section{Conclusion}
	
	Embedding is the fundamental part of deep neural networks, which could also be the bottleneck of the model strength. Building a more fine-grained representation at the very beginning could potentially guide the following networks, especially attention component to collect more important pieces. This paper presents a general yet effective architecture, subword-augmented embedding to enhance the word representation and effectively handle rare or unseen words. Experiments on five datasets from textual entailment and reading comprehension tasks demonstrate significant performance gains over the baselines. Unlike most existing works, which introduce either complex attentive architectures, handcrafted features or extra knowledge resources, our model is much more simple yet effective.
	The proposed method takes variable-length subwords segmented by unsupervised segmentation measures, without relying on any predefined linguistic resource. Thus the proposed method is also suitable for various open vocabulary NLP tasks. Our work discloses that the deep internals of sub-word level embeddings are crucial, helping downstream models to absorb different signals.


	\ifCLASSOPTIONcaptionsoff
	\newpage
	\fi

	
	
  	\bibliographystyle{plainnat}
	\bibliography{subword}
	%
	
	%
	
	\begin{IEEEbiography}[{\includegraphics[width=1in,height=1.25in,clip,keepaspectratio]{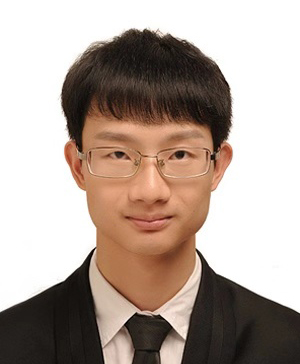}}]{Zhuosheng Zhang}
  		received the Bachelor's degree in internet of things from Wuhan University,  Wuhan, China, in 2016. He has been working toward the Master's degree in computer science and engineering with the Center for Brain-like Computing and Machine Intelligence of Shanghai Jiao Tong University, Shanghai, China. His research interests lie within deep learning for natural language processing and understanding, and he is particularly interested in question answering and machine reading comprehension. 
	\end{IEEEbiography}
 \vspace{-6mm}
	\begin{IEEEbiography}[{\includegraphics[width=1in,height=1.25in,clip,keepaspectratio]{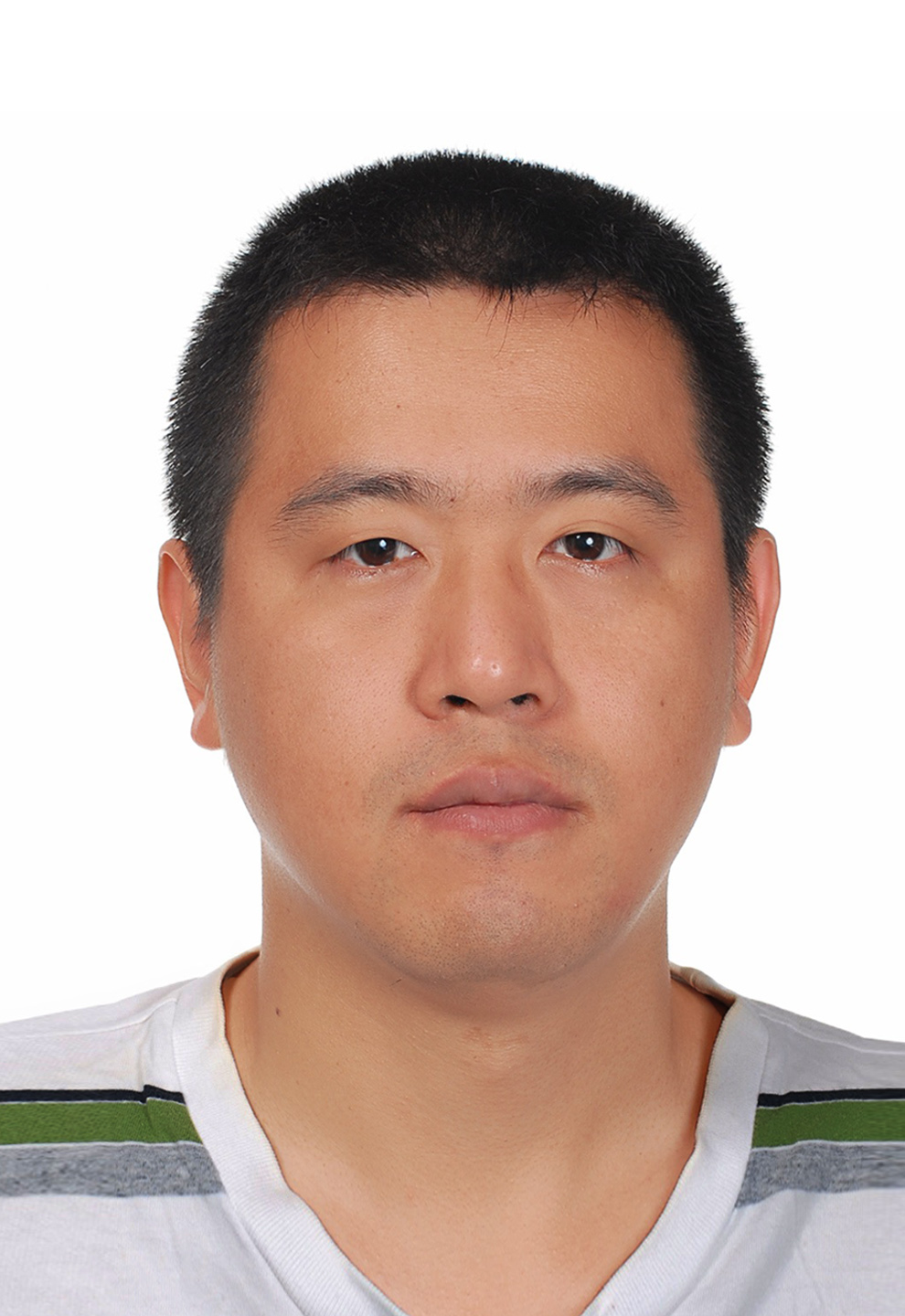}}]{Hai Zhao}
		received the BEng degree in sensor and instrument engineering, and the MPhil degree in control theory and engineering from Yanshan University in 1999 and 2000, respectively,
		and the PhD degree in computer science from Shanghai Jiao Tong University, China in 2005. 
		He is currently a full professor at department of computer science and engineering,  Shanghai Jiao Tong University after he joined the university in 2009. 
		He was a research fellow at the City University of Hong Kong from 2006 to 2009, a visiting scholar in Microsoft Research Asia in 2011, a visiting expert in NICT, Japan in 2012.
		He is an ACM professional member, and served as area co-chair in ACL 2017 on Tagging, Chunking, Syntax and Parsing, (senior) area chairs in ACL 2018, 2019 on Phonology, Morphology and Word Segmentation.
		His research interests include natural language processing and related machine learning, data mining and artificial intelligence.
	\end{IEEEbiography}
  \vspace{-8mm}
	\begin{IEEEbiography}[{\includegraphics[width=1in,height=1.25in,clip,keepaspectratio]{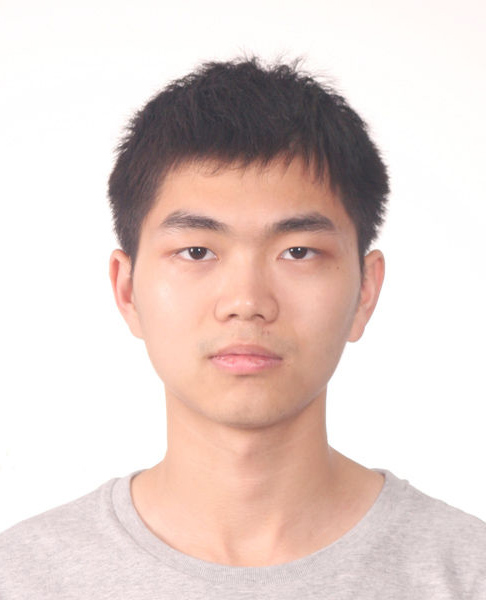}}]{Kangwei Ling}
		 received the B.S. degree from Shanghai Jiao Tong University, Shanghai, China, in 2018. He is currently pursuing the M.S. degree in Computer Science at Columbia University. During his undergraduate study, he had been doing research on natural language processing at BCMI Shanghai Jiao Tong University. His research focuses machine reading comprehension.
	\end{IEEEbiography}
	
	\vspace{-12mm}
	\begin{IEEEbiography}[{\includegraphics[width=1in,height=1.25in,clip,keepaspectratio]{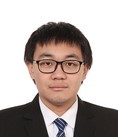}}]{Jiangtong Li}
		Jiangtong Li is an undergraduate student in Shanghai Jiao Tong University, Shanghai, China. Since 2015, he has joined the Center for Brain-like Computing and Machine Intelligence of Shanghai Jiao Tong University, Shanghai, China. His research focuses on natural language processing, especially in dialogue system.
	\end{IEEEbiography}

 \vspace{-12mm}
	\begin{IEEEbiography}[{\includegraphics[width=1in,height=1.25in,clip,keepaspectratio]{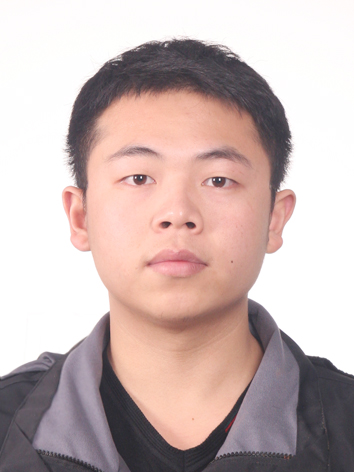}}]{Zuchao Li}
	 received the B.S. degree from Wuhan University, Wuhan, China, in 2017. Since 2017, he has been a Ph.D. student with the Center for Brain-like Computing and Machine Intelligence of Shanghai Jiao Tong University, Shanghai, China. His research focuses on natural language processing, especially syntactic and semantic parsing. 
	\end{IEEEbiography}
 \vspace{-12mm}
	\begin{IEEEbiography}[{\includegraphics[width=1in,height=1.25in,clip,keepaspectratio]{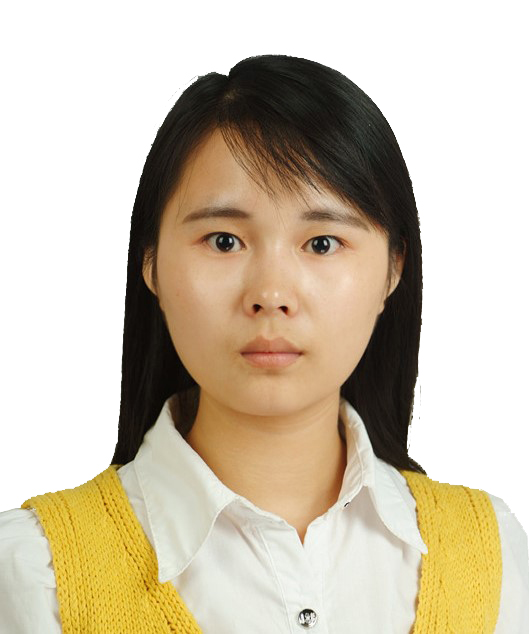}}]{Shexia He}
	received the B.S. degree from University of Electronic Science and Technology of China, in 2017. Since then, she has been a master student in Department of Computer Science and Engineering, Shanghai Jiao Tong University. Her research focuses on natural language processing, especially shallow semantic parsing, semantic role labeling.
	\end{IEEEbiography}
	 \vspace{-12mm}
	\begin{IEEEbiography}[{\includegraphics[width=1in,height=1.25in,clip,keepaspectratio]{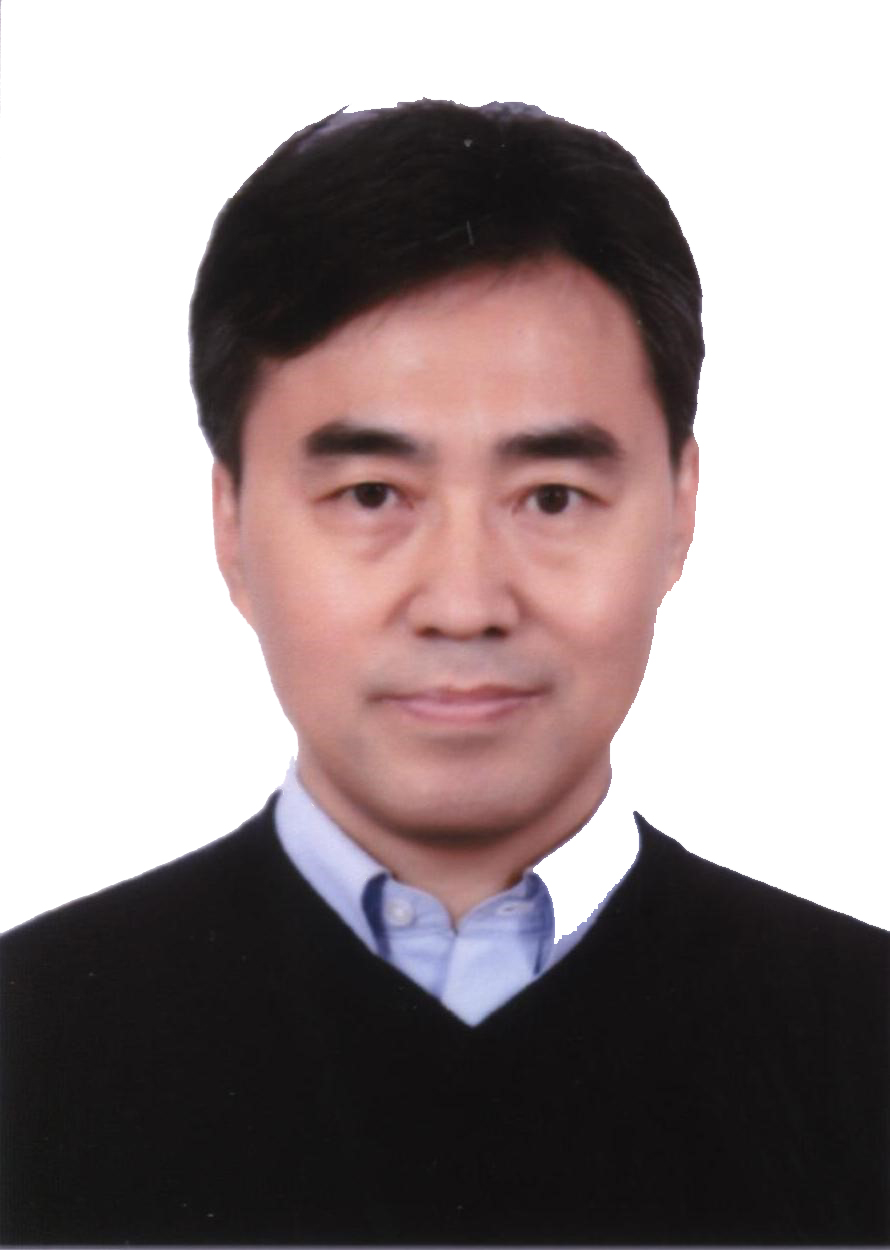}}]{Guohong Fu}
    received his Bachelor degree in automobile design, Master degree in mechanical design, and Ph.D. degree in computer science from Harbin Institute of Technology in 1990, 1993, and 2001, respectively. He is now a professor at the Soochow University. From 2001 to 2002, he worked as a researcher at InfoTalk Technology (Singapore). He was a post-doctoral fellow from 2002 to 2005 and an honorary assistant professor from 2003 to 2006 at the University of Hong Kong. From 2006 to 2007, he worked as a research fellow at the City University of Hong Kong. From 2007 to 2019, he worked as a professor at Heilongjiang University. His current research interests include natural language processing, text mining and dialogue system.
	\end{IEEEbiography}
	
	
	

\end{document}